\documentclass[11pt,a4paper]{article}
\usepackage{url}
\usepackage{hyperref}
\usepackage[hyperref]{acl2018}
\usepackage{times}
\usepackage{latexsym}
\usepackage{amsmath}
\usepackage{amssymb}
\usepackage{amsfonts}
\usepackage{bm}
\usepackage{xcolor}
\usepackage{xspace}
\usepackage{mathtools}
\usepackage{booktabs}
\usepackage{tabularx}
\usepackage{multirow}
\usepackage[toc,page]{appendix}
\usepackage{enumitem}
\usepackage{standalone}
\usepackage{dsfont}

\usepackage{tikz-dependency}
\usetikzlibrary{automata,decorations.markings,arrows,positioning,matrix,calc,patterns,angles,quotes,calc}
\usepackage[shortcuts]{extdash}  
\usepackage{pgfplots}
\usepackage{dblfloatfix}
\usepackage{float}

\aclfinalcopy 
\def\aclpaperid{946} 

\newcommand{\ra}[1]{\renewcommand{\arraystretch}{#1}}

\newcommand{\com}[1]{}
\newcommand{\resolved}[1]{}
\renewcommand{\cite}[1]{\citep{#1}}
\newcommand{\isection}[2]{\section{#1}\label{sec:#2}}
\newcommand{\isectionb}[1]{\section{#1}\label{sec:#1}}
\newcommand{\isubsection}[2]{\subsection{#1}\label{ssec:#2}}

\newcommand{\secref}[1]{Section~\ref{sec:#1}}
\newcommand{\appref}[1]{Appendix~\ref{sec:#1}}
\newcommand{\subsecref}[1]{Section~\ref{ssec:#1}}
\newcommand{\figref}[1]{Figure~\ref{#1}}
\newcommand{\tabref}[1]{Table~\ref{#1}}
\newcommand{\interalia}[1]{\citep{#1}} 
\newcommand{\softp}{pattern\xspace}
\newcommand{\softps}{patterns\xspace}
\newcommand{\SoftP}{SoPa\xspace}
\newcommand{\Softp}{SoPa}
\newcommand{\StartState}{\texttt{START}}
\newcommand{\EndState}{\texttt{END}}

\newcommand{\term}[1]{\textbf{#1}} 
\newcommand{\happy}{main path\xspace}
\newcommand{\epstrans}{$\epsilon$\=/transition} 

\newcommand{\relat}{an extension of\xspace}

\newcommand{\codeurl}{\url{https://github.com/Noahs-ARK/soft_patterns}}

\newcommand{\tensor}[1]{\mathbf{#1}}
\newcommand{\seq}[1]{\bm{#1}}
\newcommand{\R}{\mathbb{R}}

\newcommand{\camready}[1]{#1}
\DeclareMathOperator{\maxmul}{maxmul}
\DeclareMathOperator{\eps}{eps}
\definecolor{DarkGreen}{RGB}{0,111,0}
\definecolor{DarkBlue}{RGB}{0,0,111}
\definecolor{DarkRed}{RGB}{111,0,0}
\definecolor{DarkOrange}{RGB}{200,111,0}

\title{\SoftP: Bridging CNNs, RNNs, and Weighted Finite-State Machines}

\author{Roy Schwartz\thanks{~~The first two authors contributed equally.}~~$^\diamondsuit$$^\heartsuit$ \quad
  Sam Thomson\footnotemark[1]~~$^\clubsuit$  \quad
  Noah A. Smith$^\diamondsuit$ \\
  $^\diamondsuit$Paul G. Allen School of Computer Science \& Engineering,
  University of Washington \\ 
  $^\clubsuit$Language Technologies Institute,
  Carnegie Mellon University \\ 
  $^\heartsuit$Allen Institute for Artificial Intelligence \\
  {\tt \{roysch,nasmith\}@cs.washington.edu,
    sthomson@cs.cmu.edu}
}

\date{}

\begin{document}
\maketitle
\begin{abstract}
Recurrent and convolutional neural networks comprise two distinct families of
models that have proven to be useful for encoding natural language utterances.
In this paper we present \SoftP, a new model that aims to bridge these two approaches.
\SoftP combines neural representation learning with weighted finite-state automata (WFSAs) to learn a soft version of traditional surface patterns. 
We show that \SoftP is \relat a one-layer CNN, and that such CNNs are equivalent to a restricted version of \SoftP, and accordingly, to a restricted form of WFSA. 
Empirically, on three text classification tasks, \SoftP is comparable or better than both a BiLSTM (RNN) baseline and a CNN baseline, and is particularly useful in small data settings. 

\end{abstract}

\section{Introduction}

Recurrent neural networks (RNNs; \citealp{Elman:1990}) and convolutional neural networks (CNNs; \citealp{lecun_gradient-based_1998}) 
are two of the most useful text representation learners in NLP \citep{Goldberg:2016}.
These methods are generally considered to be quite different: 
the former encodes an arbitrarily long sequence of text, and is highly expressive \citep{Siegelmann:1995}.
The latter is more local, encoding fixed length windows, and accordingly less expressive.
In this paper, we seek to bridge the gap between RNNs and CNNs, presenting \term{\SoftP} (for \textbf{So}ft {\bf Pa}tterns), a model that lies in between them.

\begin{figure}[t]

\begin{center}
\begin{tikzpicture}[remember picture,minimum size=3cm,font=\Huge,->,>=stealth',auto,node distance=1cm,
  thick,main node/.style={circle,draw,font=\sffamily\Large\bfseries}]
\newcommand{\ratio}[0]{0.2}
\newcommand{\dist}[0]{.75}
\newcommand{\textsize}{\footnotesize}

\hspace{-.4cm}
\node[circle,draw,black,scale=\ratio] (h1) at (-13.5,-20) {\StartState};
\node[circle,draw,black,scale=\ratio, right = \dist cm of h1] (h2)  {1};
\node[circle,draw,black,scale=\ratio, right = \dist cm of h2] (h3)  {2};
\node[circle,draw,black,scale=\ratio, right = \dist cm of h3] (h4)  {3};
\node[circle,draw,black,scale=\ratio, right = \dist cm of h4] (h5)  {4};
\node[double,circle,draw,black,scale=\ratio, right = \dist cm of h5] (h6)  {\EndState};

 \path[every node/.style={font=\sffamily\textsize}]
    (h1) edge node [right, text width=.8cm,midway,above=-3em ] {\it What} (h2)
    (h2) edge node [right, text width=.3cm,midway,above=-3em] {\it a} (h3)
    (h3) edge node [right, text width=.9cm,midway,above=-3em ] {\it great} (h4)
    (h4) edge node [right, text width=.1cm,midway,above=-3em ] {\bf X} (h5)
    (h5) edge node [right, text width=.3cm,midway,above=-3em ] {\it !} (h6);
 \path[every node/.style={font=\sffamily\textsize}] 
 (h4) edge[loop] node[above=-1cm,align=center] {\color{DarkGreen}{\it funny}, \\ \color{DarkBlue}{\it magical}} ();
 \path[every node/.style={font=\sffamily\textsize}]
     (h2) edge[bend right] node [below=-1.2cm]{\large{\color{orange}{$\epsilon$}}} (h3);

\end{tikzpicture}
\end{center}
\setlength{\abovecaptionskip}{-1.25cm}
\setlength{\belowcaptionskip}{-0.5cm}

\caption{\label{WFSA}
A representation of a surface pattern as a six-state automaton.
\term{Self-loops} allow for repeatedly inserting words (e.g., ``{\it funny}''). 
\term{\epstrans s} allow for dropping words (e.g., ``{\it a}'').
}
\end{figure}
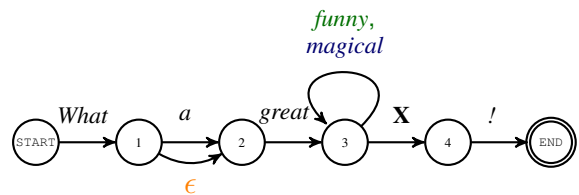

\SoftP is a neural version of a weighted finite-state automaton (WFSA), with a restricted set of transitions.
Linguistically, \SoftP is appealing as it is able to capture a soft notion of surface
patterns (e.g., \textit{``what a great \textbf{X} !''}; \citealp{Hearst:1992}), where some words may be dropped, inserted, or replaced with similar words (see \figref{WFSA}).
From a modeling perspective, \SoftP{} is interesting because WFSAs are
well-studied and come with efficient and flexible inference algorithms
\interalia{mohri_finite_1997,eisner_parameter_2002} that \SoftP{} can take
advantage of.

\SoftP defines a set of soft patterns of different lengths, with each pattern represented as a WFSA  (\secref{soft-patterns}). 
While the number and lengths of the patterns are hyperparameters, the patterns themselves are learned end-to-end.
\SoftP then represents a document with a vector that is the aggregate of the
scores computed by matching each of the patterns with each span in the
document.
Because \SoftP defines a hidden state that depends on the input token and the
previous state, it can be thought of as a simple type of RNN.

We show that \SoftP is \relat a one-layer CNN (\secref{CNN}).
Accordingly, one-layer CNNs can be viewed as a collection of linear-chain
WFSAs, each of which can only match fixed-length spans,
while our extension allows matches of flexible-length.
As a simple type of RNN that is more expressive than a CNN, \SoftP
helps to link CNNs and RNNs.

To test the utility of \SoftP, we experiment with three text classification tasks (\secref{Experiments}).
We compare against four baselines, including both a bidirectional LSTM and a CNN. 
Our model performs on par with or better than all baselines on all tasks (\secref{Results}). 
Moreover, when training with smaller datasets, \SoftP is particularly useful, outperforming all models by substantial margins.
Finally, building on the connections discovered in this paper, we offer a new, simple method to interpret \SoftP (\secref{Interpretability}).
This method applies equally well to CNNs.
We release our code at \codeurl.

\isection{Background}{background}

\paragraph{Surface patterns.} 
Patterns \cite{Hearst:1992} are particularly useful tool in NLP  \interalia{Lin:2003,Etzioni:2005,Schwartz:2015}. 
The most basic definition of a pattern is a sequence of words and
wildcards (e.g., ``\textbf{X} is a \textbf{Y}''), which can either be
manually defined or extracted from a corpus using cooccurrence statistics. 
Patterns can then be matched against a specific text span by replacing wildcards with concrete words. 

\citet{Davidov:2010} introduced a \emph{flexible} notion of patterns, which supports partial matching of the pattern with a given text by skipping some of the words in the pattern, or introducing new words. 
In their framework, when a sequence of text partially matches a  pattern, hard-coded partial scores are assigned to the pattern match.
Here, we represent patterns as WFSAs with neural weights, and support these partial matches in a soft manner.

\paragraph{WFSAs.}
We review weighted finite-state automata with
\epstrans s before we move on to our special case in \secref{soft-patterns}.
A WFSA\=/$\epsilon$ with $d$ states over a vocabulary $V$ is formally defined as a 
tuple $F = \langle \tensor{\pi}, \tensor{T}, \tensor{\eta} \rangle$, where
$\tensor{\pi} \in \R^{d}$ is an initial weight vector, $\tensor{T} : (V \cup \{\epsilon\}) \rightarrow \R^{d \times d}$ is a transition weight function, and $\tensor{\eta} \in \R^{d}$ is a final weight vector.
Given a sequence of words in the vocabulary $\seq{x} = \langle x_1, \ldots, x_n \rangle$,
the Forward algorithm \cite{baum_statistical_1966} scores $\seq{x}$ with respect to $F$.
Without \epstrans s, Forward can be written as a series of matrix multiplications:
\begin{equation}
\label{eqn:forward}
p_{\text{span}}^{\prime}(\seq{x}) =
  \tensor{\pi}^\top
  \left(\prod_{i=1}^n{\tensor{T}(x_i) }\right)
  \tensor{\eta}
\end{equation}
\epstrans s are followed without consuming a word, so Equation \ref{eqn:forward} must be
updated to reflect the possibility of following any number (zero or more) of
\epstrans s in between consuming each word:\begin{equation}
\label{eqn:forward_eps}
p_{\text{span}}(\seq{x}) =
  \tensor{\pi}^\top \tensor{T}(\epsilon)^*
    \left(\prod_{i=1}^n{
      \tensor{T}(x_i) \tensor{T}(\epsilon)^*
    }\right)
    \tensor{\eta}
\end{equation}
where $^*$ is matrix asteration: $\mathbf{A}^* \coloneqq \sum_{j=0}^\infty{\mathbf{A}^j}$.
In our experiments we use a first-order approximation,
$\mathbf{A}^* \approx \mathbf{I} + \mathbf{A}$, which corresponds to allowing zero or one \epstrans\ at a time.
When the FSA $F$ is probabilistic, the result of the Forward algorithm
can be interpreted as the marginal probability of all paths
through $F$ while consuming $\seq{x}$ (hence  the symbol ``$p$'').

The Forward algorithm can be generalized to any semiring \citep{eisner_parameter_2002}, a fact
that we make use of in our experiments and analysis.\footnote{The semiring parsing view \cite{goodman_semiring_1999} has produced unexpected connections in the past
\cite{eisner_inside_2016}.
We experiment with max-product and max-sum semirings, but note that
our model could be easily updated to use any semiring.
}
The vanilla version of Forward uses the sum-product semiring: $\oplus$
is addition, $\otimes$ is multiplication.
A special case of Forward  is the Viterbi algorithm
\cite{viterbi_error_1967}, which sets $\oplus$
to the $\max$ operator. Viterbi finds the \emph{highest scoring} path
through $F$ while consuming $\seq{x}$.
Both Forward and Viterbi have runtime $O(d^3 + d^2 n),$
requiring just a single linear pass through the phrase. 
Using first-order approximate asteration, this runtime drops to $O(d^2 n)$.\footnote{In our case, we also use a sparse transition matrix (\subsecref{soft-patterns:as_wfsas}), which further reduces our runtime to $O(dn)$.}

Finally, we note that Forward scores are for \emph{exact matches}---the entire phrase must be consumed.
We show in \subsecref{soft-patterns:encoding} how phrase-level scores can be summarized into a document-level score.

\isection{\SoftP: A Weighted Finite-State Automaton RNN}{soft-patterns}

We introduce \SoftP,  a WFSA-based RNN,
which is designed to represent text as collection of surface pattern occurrences. 
We start by showing how a single \softp can be represented as a WFSA\=/$\epsilon$ (\subsecref{soft-patterns:as_wfsas}).
Then we describe how to score a complete document using a \softp (\subsecref{soft-patterns:encoding}),
and how multiple \softps can be used to encode a document (\subsecref{soft-patterns:classifying}).
Finally, we show that \SoftP can be seen as a simple variant of an RNN (\subsecref{RNN}).

\isubsection{Patterns as WFSAs}{soft-patterns:as_wfsas}

We describe how a \softp can be represented as a WFSA\=/$\epsilon$.
We first assume a single \softp. 
A \softp \emph{is} a WFSA\=/$\epsilon$, but we impose hard constraints on its shape, and its transition weights are given by differentiable functions that have the power to capture concrete words, wildcards, and everything in between.
Our model is designed to behave similarly to  flexible hard
patterns (see \secref{background}), but to be learnable directly and ``end-to-end''
through backpropagation.
Importantly, it will still be interpretable as simple, almost linear-chain, WFSA\=/$\epsilon$.

Each pattern has a sequence of $d$ states (in our experiments we use patterns of varying lengths between 2 and 7).
Each state $i$ has exactly three possible outgoing transitions: a
\term{self-loop}, which allows the pattern to consume a word without
moving states, a \term{\happy}  transition to state $i+1$ which allows the
pattern to consume one token and move forward one state, and an
\term{\epstrans} to state $i+1$, which allows the pattern to move
forward one state without consuming a token.
All other transitions are given score 0.
When processing a sequence of text with a pattern $p$, we start with a special \StartState\ state, and only move forward (or stay put), until we reach the special \EndState\ state.\footnote{To ensure that we start in the \StartState\ state and end in the \EndState\ state, we fix $\pi=[1,0,\dots,0]$ and $\eta=[0,\dots,0,1]$.}
A pattern with $d$ states will tend to match token spans of length $d - 1$ (but possibly shorter spans due to \epstrans s, or longer spans due to self-loops).
See \figref{WFSA} for an illustration.

Our transition function, $\tensor{T}$, is a parameterized function that returns a $d \times d$ matrix.
For a word $x$:\begin{equation}
\label{eqn:main-self-transitions}
\left[\tensor{T}(x)\right]_{i,j} =
  \begin{cases}
    E(\tensor{u}_{i} \cdot \tensor{v}_x + a_{i}), & \text{ if } j = i \textit{ (self-loop)} \\
    E(\tensor{w}_{i} \cdot \tensor{v}_x + b_{i}), & \text{ if } j = i+1 \\ 
    0, & \text{ otherwise,}
  \end{cases}
\end{equation}
where $\tensor{u}_{i}$ and $\tensor{w}_{i}$ are vectors of parameters, $a_{i}$  and $b_i$ are scalar parameters, $\tensor{v}_x$ is a fixed pre-trained word vector for $x$,\footnote{We use GloVe 300d 840B \cite{Pennington:2014}.} and $E$ is an encoding function, typically the identity function or sigmoid.
\epstrans s are also parameterized, but don't consume a token and depend only on the current state:\begin{equation}
\label{eqn:eps_transitions}
\left[\tensor{T}(\epsilon)\right]_{i,j} =
  \begin{cases}
    E(c_{i}), & \text{ if } j = i+1 \\
    0, & \text{ otherwise,}
  \end{cases}
\end{equation}
where $c_i$ is a scalar parameter.\footnote{Adding \epstrans s to WFSAs does not increase their expressive power, and in fact slightly complicates the Forward equations.
We use them as they require fewer parameters, and make the modeling connection between (hard) flexible patterns and our (soft) \softps more direct and intuitive.
}
As we have only three non-zero diagonals in total, the matrix multiplications in Equation~\ref{eqn:forward_eps} can be implemented using vector operations, and the overall runtimes of Forward and Viterbi are reduced to $O(dn)$.\footnote{Our implementation is optimized to run on GPUs, so the observed runtime is even {\it sublinear} in $d$.}

\paragraph{Words vs.~wildcards.}
Traditional  {\it hard} patterns distinguish between words and wildcards. 
Our model does not explicitly capture the notion of either, but the
transition weight function can be interpreted in those terms.
Each transition is a logistic regression over the next word vector $\tensor{v}_x$.
For example, for a \happy out of state $i$, $\tensor{T}$ has two parameters, $\tensor{w}_{i}$ and $b_{i}$.
If $\tensor{w}_{i}$ has large magnitude and is close to the word vector for some word $y$ (e.g., $\tensor{w}_i \approx 100 \tensor{v}_y$), and $b_i$ is a large negative bias (e.g., $b_i \approx -100$), then the transition is essentially matching the specific word $y$.
Whereas if $\tensor{w}_i$ has small magnitude ($\tensor{w}_i \approx
\tensor{0}$) and $b_i$ is a large positive bias (e.g., $b_i \approx
100$), then the transition is ignoring the current token and matching
 a wildcard.\footnote{A large bias increases the
eagerness to match \emph{any} word.}
The transition could also be something in between, for instance by
focusing on specific dimensions of a word's meaning encoded in the
vector, such as POS or semantic features like animacy or
concreteness \cite{\camready{Rubinstein:2015,}Tsvetkov:2015}.

\isubsection{Scoring Documents}{soft-patterns:encoding}

So far we described how to calculate how well a pattern matches a token span exactly (consuming the whole span).
To score a complete document, we prefer a score that aggregates over all matches on subspans of the document (similar to ``search'' instead of ``match'' in regular expression parlance).
We still assume a single \softp.

Either the Forward algorithm can be used to calculate  the expected
count of the pattern in the document,
$\sum_{1\leq i \leq j \leq
  n}{p_{\text{span}}(\seq{x}_{i:j})}$, or Viterbi to calculate
$s_{\text{doc}}(\seq{x})$ = $\max_{1\leq i \leq j \leq
  n}{s_{\text{span}}(\seq{x}_{i:j})}$, the score of
  the highest-scoring match.
In short documents, we expect patterns to typically occur at most once, 
so in our experiments we choose the Viterbi algorithm, i.e., the max-product semiring.

\paragraph{Implementation details.}
We give the specific recurrences we use to score documents in a single pass with this model.
We define:
\begin{equation}
[\maxmul(\tensor{A}, \tensor{B})]_{i,j} =
  \max_{k}{
    A_{i,k} B_{k,j}
  }.
\end{equation}
We also define the following for taking zero or one \epstrans s:
\begin{equation}
\eps{(\tensor{h})} =
  \maxmul{\left(
    \tensor{h},
    \max(\tensor{I}, \tensor{T}(\epsilon))
  \right)}
\end{equation}
where $\max$ is element-wise max.
We maintain a row vector $\tensor{h}_t$ at each token:\footnote{%
Here a row vector $\tensor{h}$ of size $n$ can also be viewed as a $1 \times n$ matrix.}
\begin{subequations}
\begin{align}
\label{eqn:state_eqns:start}
\tensor{h}_0 = &
  \eps(\tensor{\pi^\top}), \\
\label{eqn:state_eqns:next}
\tensor{h}_{t+1} = &
  \max{\left(
    \eps(\maxmul{(\tensor{h}_t, \tensor{T}(x_{t+1}))}),
    \tensor{h}_0
  \right)},
\end{align}
\end{subequations}
and then extract and aggregate \EndState\ state values:
\begin{subequations}
\begin{align}
\label{eqn:state_eqns:score_t}
s_t = &
  \maxmul{(\tensor{h}_t, \tensor{\eta})}, \\
\label{eqn:state_eqns:score_sum}
s_{\text{doc}} = &
  \max_{1 \leq t \leq n}{s_t}.
\end{align}
\end{subequations}
$[\tensor{h}_t]_i$ represents the score of the best path through the pattern that ends in state $i$ after consuming $t$ tokens.
By including $\tensor{h}_0$ in Equation~\ref{eqn:state_eqns:next}, we are accounting for spans that start at time $t+1$.
$s_t$ is the maximum of the exact match scores for all spans ending at token $t$.
And $s_{\text{doc}}$ is the maximum score of any subspan in the document.

\isubsection{Aggregating Multiple Patterns}{soft-patterns:classifying}

We describe how $k$ patterns are aggregated to score a document.
These $k$ patterns give $k$ different $s_{\text{doc}}$ scores for the document,
which are stacked into a vector $\tensor{z} \in \R^k$ and constitute the final document representation of \SoftP. 
This vector representation can be viewed as a feature vector.
In this paper, we feed it into a multilayer perceptron (MLP), culminating in a softmax to give a probability distribution over document labels.
We minimize cross-entropy, allowing the \SoftP and MLP parameters to be learned end-to-end.

\SoftP uses a total of $(2 e + 3) d k$ parameters,
where $e$ is the word embedding dimension, $d$ is the number of states and $k$ is the
number of patterns.
For comparison, an LSTM with a hidden dimension of $h$ has
$4 ((e + 1) h + h^2)$.
In \secref{Results} we show that \SoftP consistently uses fewer parameters than a BiLSTM baseline to
achieve its best result.

\begin{figure*}[t]
\begin{center}
\hspace{-0.3cm}
  \includegraphics[trim={0 0.1cm 0 0.18cm},clip,width=.9\textwidth]{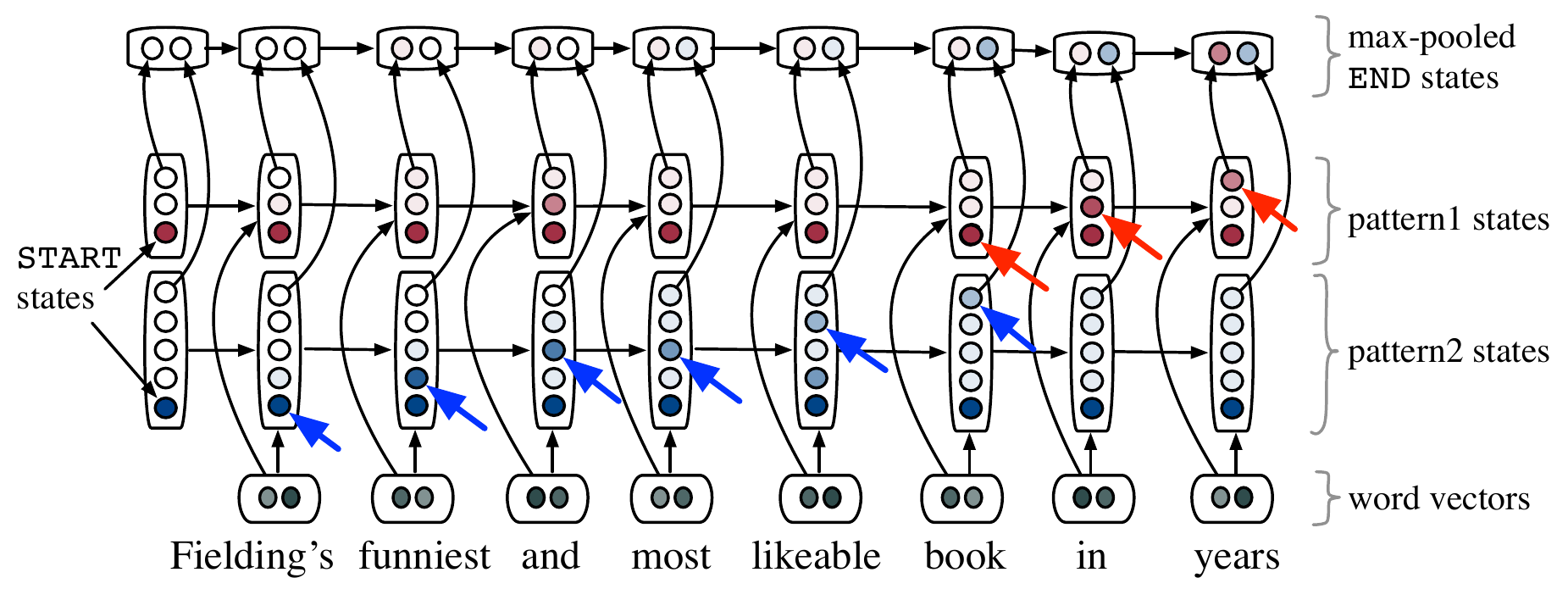}
\end{center}

\caption{
  \label{fig:model}
  State activations of two patterns as they score a document.
  pattern1 (length three) matches on \textit{``in years''}.
  pattern2  (length five) matches on \textit{``funniest and most likeable book''}, using a self-loop to consume the token \textit{``most''}.
  Active states in the best match are marked with arrow cursors.
}
\end{figure*}

\isubsection{\SoftP as an RNN}{RNN}
\SoftP can be considered an RNN.
As shown in \subsecref{soft-patterns:encoding}, a single pattern with $d$ states has a hidden state vector of size $d$. 
Stacking the $k$ hidden state vectors of $k$ patterns into one vector of size $k \times d$ can be thought of as the hidden state of our model.
This hidden state is, like in any other RNN, dependent of the input and the previous state.
Using self-loops, the hidden state at time point $i$ can in theory depend on the entire history of tokens up to $x_i$ (see \figref{fig:model} for illustration).
We do want to discourage the model from following too many self-loops, only doing so if it results in a better fit with the remainder of the pattern.
To do this we use the sigmoid function as our encoding function $E$ (see Equation~\ref{eqn:main-self-transitions}), which means that all transitions have scores strictly less than 1.
This works to keep pattern matches close to their intended length.
Using other encoders, such as the identity function, can result in different dynamics, potentially encouraging rather than discouraging self-loops.

Although even single-layer RNNs are Turing complete~\cite{Siegelmann:1995},
\SoftP's expressive power depends on the semiring.
When a WFSA is thought of as a function from finite sequences
of tokens to semiring values, it is restricted to the class of functions known as
\term{rational series}~\cite{schutzenberger_definition_1961,droste_kleeneschutzenberger_1999,sakarovitch_rational_2009}.\footnote{Rational series generalize recognizers of \emph{regular languages}, which are
the special case of the Boolean semiring.
}
It is unclear how limiting this theoretical restriction is in practice, especially when \SoftP
is used as a component in a larger network.
We defer the investigation of the exact computational properties of \SoftP to future work.
In the next section, we show that \SoftP is \relat a one-layer CNN, and hence more expressive.

\isection{\SoftP as a CNN Extension}{CNN}
A convolutional neural network \citep[\term{CNN};][]{lecun_gradient-based_1998} moves a fixed-size sliding window over the document, producing a vector representation for each window.
These representations are then often summed, averaged, or max-pooled to produce a document-level representation \interalia{kim_convolutional_2014,yin_multichannel_2015}.
In this section, we show that \SoftP is \relat one-layer, max-pooled CNNs.

To recover a CNN from a soft pattern with $d+1$ states, we first remove
self-loops and \epstrans s, retaining only the \happy transitions.
We also use the identity function as our encoder $E$
(Equation~\ref{eqn:main-self-transitions}), and use the max-sum semiring.
With only main path transitions, the network will 
not match any span that is not exactly $d$ tokens long.
Using max-sum, spans of length $d$ will be assigned the score:
\begin{subequations}
\begin{align}
\label{eqn:cnn-span}
  s_{\text{span}}(\seq{x}_{i:i+d})
    =& \sum_{j=0}^{d-1}{\tensor{w}_j \cdot \tensor{v}_{x_{i+j}} + b_j}, \\
    =& \tensor{w}_{0:d} \cdot \tensor{v}_{x_{i:i+d}}
      + \sum_{j=0}^{d-1}{b_j},
\end{align}
\end{subequations}
where $\tensor{w}_{0:d} = [ \tensor{w}_0^\top; \ldots ; \tensor{w}_{d-1}^\top ]^\top$,  $\tensor{v}_{x_{i:i+d}} = [ \tensor{v}_{x_i}^\top; \ldots; \tensor{v}_{x_{i+d-1}}^\top ]^\top$.
Rearranged this way, we recognize the span score as an affine transformation
of the concatenated word vectors $\tensor{v}_{x_{i:i+d}}$.
If we use $k$ patterns, then together their span scores correspond to a linear
filter with window size $d$ and output dimension $k$.\footnote{This variant of \SoftP has $d$ bias parameters, which correspond to only a single bias parameter in a CNN.
The redundant biases may affect optimization but are an otherwise unimportant difference.}
A single pattern's score for a document is:
\begin{equation}
\label{eqn:cnn-doc}
  s_\text{doc}(\seq{x}) = \max_{1 \leq i \leq n-d+1}{s_\text{span}(\seq{x}_{i:i+d})}.
\end{equation}
The $\max{}$ in Equation~\ref{eqn:cnn-doc} is calculated for each pattern independently, corresponding exactly to element-wise max-pooling of the CNN's output layer.

Based on the equivalence between this impoverished version of \SoftP and CNNs,
we conclude that one-layer CNNs are learning an even more restricted class of
WFSAs (linear-chain WFSAs) that capture only fixed-length \softps.

One notable difference between \SoftP and arbitrary CNNs is that in general CNNs can use
any filter (like an MLP over $\tensor{v}_{x_{i:i+d}}$, for example).
In contrast, in order to efficiently pool over flexible-length spans, \SoftP is restricted to operations that follow the semiring laws.\footnote{The max-sum semiring corresponds to a linear filter with max-pooling.
Other semirings could potentially model more interesting interactions, but we leave this to future work.}

As a model that is more flexible than a one-layer CNN, but (arguably) less expressive than
many RNNs, \SoftP lies somewhere on the continuum between these two approaches.
Continuing to study the bridge between CNNs and RNNs is an exciting direction for future research.

\isectionb{Experiments}

To evaluate \SoftP, we  apply it to text classification tasks. 
Below we describe our datasets and baselines.
More details can be found in \appref{Experimental}.

\paragraph{Datasets.}
We experiment with three binary classification datasets. 

\begin{itemize}[itemsep=0pt,topsep=3pt,leftmargin=*]
\item {\bf SST}. The Stanford Sentiment Treebank \cite{Socher:2013}\camready{\footnote{\url{https://nlp.stanford.edu/sentiment/index.html}}} contains roughly 10K movie reviews from Rotten Tomatoes,\camready{\footnote{\url{http://www.rottentomatoes.com}}} labeled on a scale of 1--5.
We consider the binary task, which considers 1 and 2 as
negative, and 4 and 5 as positive (ignoring 3s).
It is worth noting that this dataset also contains
syntactic phrase level annotations, providing a sentiment label to parts of sentences. 
In order to experiment in a realistic setup, we only consider the complete sentences,
and ignore syntactic annotations at train or test time.
The number of training/development/test sentences in the dataset is 6,920/872/1,821.

\item{\bf Amazon}. The Amazon Review Corpus \cite{McAuley:2013}\camready{\footnote{\url{http://riejohnson.com/cnn_data.html}}} contains electronics product reviews\camready{,
a subset of a larger review dataset}.
Each document in the dataset contains a review and a summary. 
Following \citet{Yogatama:2015}, we only use the reviews part, focusing on positive and negative reviews.
The number of training/development/test samples is 20K/5K/25K.

\item{\bf ROC}. The ROC story cloze task \cite{Mostafazadeh:2016} is a story understanding task.\camready{\footnote{\url{http://cs.rochester.edu/nlp/rocstories/}}}
The task is composed of four-sentence story prefixes, followed by two competing endings: one that makes the joint five-sentence story coherent, and another that makes it incoherent. 
Following \citet{Schwartz:2017}, we treat it as a style detection task: 
we treat all ``right'' endings as positive samples and all ``wrong'' ones as negative, and we ignore the story prefix. 
We split the development set into train and development (of sizes 3,366 and 374 sentences, respectively), and take the test set as-is (3,742 sentences).
\end{itemize}

\paragraph{Reduced training data.}
In order to test our model's ability to learn from small datasets, we also randomly sample 100, 500, 1,000 and 2,500 SST training instances and 
 100, 500, 1,000, 2,500, 5,000, and 10,000 Amazon training instances.
Development and test sets remain the same.

\paragraph{Baselines.}
We compare to four baselines: 
a BiLSTM, a one-layer CNN, DAN (a simple alternative to RNNs) and a feature-based classifier trained with hard-pattern features.

\begin{itemize}[itemsep=0pt,topsep=3pt,leftmargin=*]
\item{\bf BiLSTM.} 
Bidirectional LSTMs have been successfully used in the past for text classification tasks \cite{Zhou:2016}.
We learn a one-layer BiLSTM representation of the
document, and feed the average of all hidden states to an MLP. 

\item{\bf CNN}.
CNNs are particularly useful for text classification \cite{kim_convolutional_2014}.
We train a one-layer CNN with max-pooling, and feed the resulting representation to an MLP.

\item{\bf DAN}. 
We learn a deep averaging network with word dropout \cite{Iyyer:2015}, a simple but strong text-classification baseline.

\item{\bf Hard}. 
We train a logistic regression classifier with hard-pattern features. 
Following \citet{Tsur:2010}, we replace low frequency words with a special wildcard symbol.
We learn sequences of 1--6 concrete words,
where any number of wildcards can come between two adjacent 
words. \camready{We consider words occurring with frequency of at least 0.01\% of our training set as concrete words, and words occurring in frequency 1\% or less as wildcards.\footnote{Some words may serve as both words and wildcards. See \citet{Davidov:2008} for discussion.}}

\end{itemize}

\paragraph{Number of patterns.}

\SoftP requires specifying the number of patterns to be learned, and their lengths.
Preliminary experiments showed that the model doesn't benefit from more than a few dozen patterns.
We experiment with several configurations of patterns of different lengths, generally considering 0, 10 or 20 patterns of each pattern length between 2--7.
The total number of patterns learned ranges between 30--70.\footnote{The number of patterns and their length are hyperparameters tuned on the development data (see \appref{Experimental}).}

\isectionb{Results}

\tabref{tab:results} shows our main experimental results. 
In two of the cases (SST and ROC), \SoftP outperforms all models.
On Amazon, \SoftP performs within 0.3 points of CNN and BiLSTM, and outperforms the other two baselines.
The table also shows the number of parameters used by each model for each task. 
Given enough data, models with more parameters should be expected to perform better.
However, \SoftP performs better or roughly the same as a BiLSTM, which has  $3$--$6$ times as many parameters.

\begin{table}[!t]

\setlength{\tabcolsep}{1.7pt}
\small
\centering
\begin{tabularx}{\linewidth}{@{}l X X X  @{}}
  \toprule
  {\bf Model} & {\bf ROC} & {\bf SST} & {\bf Amazon}   \\
  \midrule
  {\bf Hard} & 62.2 (4K) & 75.5 (6K) & 88.5 (67K)   \\
  {\bf DAN} & 64.3 (91K)& 83.1 (91K) & 85.4 (91K) \\
  {\bf BiLSTM} & 65.2 (844K)& 84.8 (1.5M) & {\bf 90.8} (844K) \\
  {\bf CNN} & 64.3 (155K) & 82.2  (62K) & 90.2 (305K) \\
  \midrule
  {\bf \SoftP} & {\bf 66.5} (255K)& {\bf 85.6} (255K) & 90.5 (256K)  \\
  \midrule
    {\bf \Softp$_{ms_\mathds{1}}$} & 64.4 & 84.8 & 90.0 \\
  {\bf \Softp$_{ms_\mathds{1}}\setminus$$\{sl\}$} & 63.2 & 84.6 & 89.8  \\
  {\bf \Softp$_{ms_\mathds{1}}\setminus$$\{\epsilon\}$} & 64.3 & 83.6 & 89.7 \\
  {\bf \Softp$_{ms_\mathds{1}}\setminus$$\{sl,\epsilon\}$} & 64.0 & 85.0 & 89.5\\

  \bottomrule
\end{tabularx}
\caption{\label{tab:results} Test classification accuracy (and the number of parameters used).
The bottom part shows our ablation results:  \SoftP{}: our full model. 
\Softp$_{ms_\mathds{1}}$: running with max-sum semiring (rather than max-product), with the identity function as our encoder $E$ (see Equation~\ref{eqn:main-self-transitions}).
$sl$: self-loops, $\epsilon$: $\epsilon$ transitions.
The final row is equivalent to a one-layer CNN.}
\end{table}

Figure \ref{fig:training-size} shows a comparison of all models on the SST and Amazon 
datasets with varying training set sizes. 
\SoftP is substantially outperforming all baselines, in particular BiLSTM, on small datasets (100 samples).
This suggests that \SoftP is better fit to learn from small datasets.

\pgfplotsset{every tick label/.append style={font=\small}}

\newcommand{\figscale}{0.5}
\newcommand{\legendtextsize}{12}
\begin{figure}[!t]
\begin{center}
\begin{minipage}{.2\textwidth}
\begin{tikzpicture}[scale=\figscale]
\begin{axis}[     
y label style={at={(0.06,0.5)}},
	xmode=log,
    log ticks with fixed point,
    legend cell align={left},
    xlabel={Num.~Training Samples (SST)},
    ylabel={Classification Accuracy},
    label style={font=\Large},
    every axis plot/.append style={thick}]
    xlabel={Num.~Training Samples (SST)},
    \addplot coordinates {
        (100,75.8)
	(500,80.5)
	(1000,81.3)
	(2500,82.9)
	(6920,85.557)
    };
    \addlegendentry{{\bf \SoftP (ours)}}
    \addplot coordinates {
        (100,72)
	(500,79.4)
	(1000,80.3)
	(2500,81.9)
	(6920,83.1)
    };
    \addlegendentry{{\bf DAN}}
    \addplot coordinates {
        (100,57.8)
	(500,61.7)
	(1000,68.5)
	(2500,71.1)
	(6920,75.5)
    };
    \addlegendentry{{\bf Hard}}
    \addplot coordinates {
        (100,71.5)
	(500,79.4)
	(1000,81.4)
	(2500,83.4)
	(6920,84.8)
    };
        \addlegendentry{{\bf BiLSTM}}
    \addplot[color=DarkGreen] coordinates {
        (100,73.2)
	(500,79.6)
	(1000,81.7)
	(2500,84.1)
	(6920,82.2)
    };
        \addlegendentry{{\bf CNN}}
        
          \legend{};

\end{axis}

\end{tikzpicture}
\end{minipage}
\qquad
\begin{minipage}{.2\textwidth}
\begin{tikzpicture}[scale=\figscale]
\begin{axis}[
	xmode=log,
    log ticks with fixed point,
    xlabel={Num.~Training Samples (Amazon)},   
    legend cell align={left},
    label style={font=\Large},
    legend style={
		at={(1.,0.490),anchor=west, font=\fontsize{\legendtextsize}{5}\selectfont}
        },
    every axis plot/.append style={thick}]
    xlabel={Num.~Training Samples},
    \addplot coordinates {
        (100,77.2)
	(500,82)
	(1000,84.2)
	(2500,86.3)
	(5000, 88)
	(10000, 88.8)
	(20000,90.5)
    };
    \addlegendentry{\bf \SoftP (ours)}
    \addplot coordinates {
	(100,71.4)
	(500,78.7)
	(1000,79.7)
	(2500, 82.1)
	(5000, 83.7)
	(10000,84.5)
	(20000,85.4)
    };
    \addlegendentry{{\bf DAN}}
    \addplot coordinates {
        (100,68.5)
	(500,78.4)
	(1000,80.1)
	(2500,83.6)
	(5000, 86)
	(10000, 87.3)
	(20000,88.5)
    };
    \addlegendentry{{\bf Hard}}
    \addplot coordinates {
        (100,69.6)
	(500,79.6)
	(1000,83.6)
	(2500,86.3)
	(5000, 88)
	(10000, 89.4)
	(20000,90.8)
    };
        \addlegendentry{{\bf BiLSTM}}

    \addplot[color=DarkGreen] coordinates {
        (100,73.9)
	(500,80.2)
	(1000,82.9)
	(2500,86.1)
	(5000, 87.5)
	(10000, 88.8)
	(20000,90.2)
    };
        \addlegendentry{{\bf CNN}}

\end{axis}

\end{tikzpicture}
\end{minipage}

\caption{\label{fig:training-size} Test accuracy on SST and Amazon with varying number of training instances.} 
\end{center}

\end{figure}

\paragraph{Ablation analysis.}

\tabref{tab:results} also shows an ablation of the differences between \SoftP and CNN:
max-product semiring with sigmoid vs.~max-sum semiring with identity, self-loops, and \epstrans s.
The last line is equivalent to a CNN with multiple window sizes.
Interestingly, the most notable difference between \SoftP and CNN is the semiring and encoder function, while $\epsilon$ transitions and self-loops have little effect on performance.\footnote{Although \SoftP does make use of them---see \secref{Interpretability}.}

\isectionb{Interpretability}

We turn to another key aspect of \SoftP---its interpretability. 
We start by demonstrating how we interpret a single \softp, and then  describe
how to interpret the decisions made by downstream classifiers that
rely on \SoftP---in this case, a sentence classifier.
Importantly, these visualization techniques are equally applicable to CNNs.

 \paragraph{Interpreting a single pattern.}
In order to visualize a pattern, we compute the pattern matching scores with each phrase in our training dataset, 
and select the $k$ phrases with the highest scores.
\tabref{patt-interpretability} shows examples of six patterns learned using the best \SoftP model on the SST dataset,
as represented by their five highest scoring phrases in the training set.
A few interesting trends can be observed from these examples. 
First, it seems our patterns encode semantically coherent expressions.
A large portion of them correspond to sentiment (the five top examples in the table),
but others capture different semantics, e.g., time expressions.

\begin{table}[t]
\small
\newcommand{\ep}{{\color{red}{$\epsilon$}} }
\newcommand{\sloop}[1]{{\color{blue}{#1$_{\textit{SL}}$}} }
\setlength{\tabcolsep}{1.5pt}
\scriptsize
\begin{tabularx}{\linewidth}{c |lllll}
  \toprule
   & \multicolumn{5}{c}{\bf Highest Scoring Phrases} \\
  \midrule
  \multirow{5}{*}{Patt.~1} 
  & thoughtful &     ,        &       reverent  &      portrait  &      of  \\
         & and  &           astonishingly&   articulate&      cast&            of     \\
         &  entertaining&    ,          &     thought-provoking& film&            with       \\ 
         & gentle&        ,   &           mesmerizing&  portrait&  of \\
         & poignant&        and&             uplifting&       story&           in \\

  \midrule

  \multirow{5}{*}{Patt.~2}  
  & 's &           \ep &                uninspired &      story &           .  \\
& this&           \ep&                bad&            on&             purpose \\ 
&this&           \ep&                 leaden&          comedy&          .      \\
& a&              \ep&                 half-assed&      film&            .         \\ 
& is&             \ep&                 clumsy         \sloop{,}&               the&             writing      \\

  \midrule

 \multirow{5}{*}{Patt.~3} 
 & mesmerizing&     portrait    &    of&              a     &         \\ 
& engrossing      &portrait    &    of    &          a     &         \\
 & clear-eyed      &portrait     &   of     &         an&             \\
 & fascinating    & portrait       & of        &      a&              \\ 
 & self-assured     & portrait&        of&              small& \\
  \midrule

\multirow{5}{*}{Patt.~4}  & honest&          ,             & and&            enjoyable& \\
& soulful&        ,         \sloop{scathing} &      and        &    joyous& \\
& unpretentious&  ,          \sloop{charming} &      ,   &           quirky &        \\
& forceful&  ,          &     and&  beautifully&\\
& energetic& ,          &     and&            surprisingly& \\

  \midrule

   \multirow{5}{*}{Patt.~5} & is &             deadly&          dull&& \\
 & a&               numbingly&       dull        &&   \\
&  is&             remarkably & dull&& \\
& is   &           a             &  phlegmatic&& \\
 & an   &           utterly&         incompetent&& \\

  \midrule

  \multirow{5}{*}{Patt.~6} & five & minutes&&& \\
  & four &minutes&&& \\
  &  final &          minutes&&& \\
  & first& half-hour&&& \\
    & fifteen&         minutes&&& \\

  \bottomrule
\end{tabularx}
\caption{\label{patt-interpretability}
Six patterns of different lengths learned by \SoftP on SST.
Each group represents a single pattern $p$, and shows the five phrases in
the training data that have the highest score for $p$.
Columns represent pattern states.
Words marked with \sloop{}are self-loops.
\ep{}symbols indicate \epstrans s.
All other words are from \happy\ transitions.}
\end{table}

Second, it seems our patterns are relatively soft, and  allow lexical flexibility. 
While some patterns do seem to fix specific words, e.g., ``of'' in the first example or ``minutes'' in the last one, 
even in those cases some of the top matching spans replace these words with other, similar words (``with'' and ``half-hour'', respectively).
Encouraging \SoftP to have more concrete words, e.g., by jointly learning the word vectors, might make \SoftP useful in other contexts, particularly as a decoder.
We defer this direction to future work.

Finally, \SoftP makes limited but non-negligible use of self-loops and epsilon steps.
Interestingly, the second example shows that one of the patterns had an \epstrans\ at the same place in every phrase.
This demonstrates a different function of \epstrans s than originally designed---they allow a \softp to 
effectively shorten itself, by learning a high \epstrans\ parameter for a certain state.

 \paragraph{Interpreting a document.}
\SoftP provides an interpretable representation of a document---a vector of the maximal matching score of each pattern with any span in the document.
To visualize the decisions of our model for a given document, we can observe the patterns and corresponding phrases that score highly within it.

To understand which of the $k$ patterns contributes most to the classification decision, we apply a leave-one-out method.
We run the forward method of the MLP layer in \SoftP $k$ times, each time zeroing-out the score of a different pattern $p$.
The difference between the resulting score and the original model score is considered $p$'s contribution.
We then consider the highest contributing patterns, and attach each one with its highest scoring phrase in that document.
Table~\ref{table:examples} shows example texts along with their most
positive and negative contributing phrases.

\newcommand{\positive}[1]{{\textcolor{DarkGreen}{\textbf{#1}}}}
\newcommand{\negative}[1]{{\textcolor{orange}{\textit{#1}}}}

\begin{table}[t]
\small

\begin{tabularx}{\linewidth}{@{}p{\columnwidth}@{}}
\toprule

\textbf{Analyzed Documents} \\
\midrule[\cmidrulewidth]
\negative{it 's dumb ,} \positive{but more importantly} , \negative{it 's just not scary}\\
\midrule[\cmidrulewidth]

\positive{}though moonlight mile is replete with \positive{acclaimed actors and actresses} and tackles a subject that 's \positive{potentially moving} , the movie is \negative{too predictable} and \negative{too self-conscious to reach a} level of \positive{high drama} \\
\midrule[\cmidrulewidth]
\positive{}While \positive{its careful pace and} seemingly \negative{opaque story} may not satisfy every moviegoer 's appetite, the film 's final scene is \positive{soaringly , transparently moving} \\
\midrule[\cmidrulewidth]
\positive{unlike the speedy wham-bam} effect \negative{of most hollywood offerings} , character development -- and more \positive{importantly, character empathy} -- \positive{is at the heart of} italian for beginners~. \\
\midrule[\cmidrulewidth]
\positive{the band 's courage in} the face of official repression \positive{is inspiring} , \positive{especially for} aging \negative{hippies} ( this one included )~. \\
\bottomrule
\end{tabularx}
\caption{\label{table:examples}
Documents from the SST training data.
Phrases with the largest contribution toward a positive
sentiment classification are in {\positive{bold green}}, and
the most negative phrases are in {\negative{italic orange}}.
}
\end{table}

\isection{Related Work}{related}

\paragraph{Weighted finite-state automata.} WFSAs and hidden Markov models\footnote{HMMs are a special case of  WFSAs \cite{Mohri:2002}.} were once popular in
automatic speech recognition \interalia{hetherington_mit_2004,moore_juicer_2006,hoffmeister_wfst_2012} and
remain popular in morphology \interalia{dreyer_non-parametric_2011,cotterell_modeling_2015}.
Most closely related to this work, neural networks have been combined with
weighted finite-state transducers to do morphological
reinflection~\cite{rastogi_weighting_2016}.
These prior works learn a single FSA or FST, whereas our model
learns a collection of simple but complementary FSAs, together encoding a
sequence.
We are the first to incorporate neural networks both
\emph{before} WFSAs (in their transition scoring functions), and \emph{after}
(in the function that turns their vector of scores into a final prediction),
to produce an expressive model that remains interpretable.

\paragraph{Recurrent neural networks.}

The ability of RNNs to represent arbitrarily long sequences of
embedded tokens has made them attractive to NLP researchers. 
The most notable variants, the long short-term memory (LSTM; \citealp{Hochreiter:1997}) and gated recurrent units (GRU; \citealp{Cho:2014}),
have become ubiquitous in NLP algorithms \citep{Goldberg:2016}.
Recently, several works introduced simpler versions of RNNs, such as recurrent additive networks \citep{Lee:2017} and Quasi-RNNs \citep{Bradbury:2017}.
Like \SoftP, these models can be seen as points along the bridge between RNNs and CNNs.

Other works have studied the expressive power of RNNs, in particular in the context of WFSAs or HMMs \cite{Cleeremans:1989,Giles:1992,Visser:2001,Chen:2018}.
In this work we relate CNNs to WFSAs, showing that a one-layer CNN with
max-pooling can be simulated by a collection of linear-chain WFSAs.

\paragraph{Convolutional neural networks.}

CNNs are prominent feature extractors in NLP, both for generating character-based embeddings \cite{Kim:2016}, and as sentence encoders for tasks like text classification \cite{yin_multichannel_2015} and  machine translation \cite{Gehring:2017}.
Similarly to \SoftP, several recently introduced variants of CNNs support varying window sizes
by either allowing several fixed window sizes \citep{yin_multichannel_2015} or by supporting non-consecutive $n$-gram matching \cite{Lei:2015,Nguyen:2016}.

\paragraph{Neural networks and patterns.}
Some works used patterns as part of a neural network. 
\citet{Schwartz:2016} used pattern contexts for
estimating word embeddings, showing improved word similarity results
compared to bag-of-word contexts.
\citet{Shwartz:2016} designed an LSTM representation for dependency patterns, using them to detect hypernymy relations. 
Here, we learn patterns as a neural version of WFSAs.

\paragraph{Interpretability.}
There have been several efforts to interpret neural models. 
The weights of the attention mechanism \cite{Bahdanau:2015} are often used to display the words that are most significant for making a prediction.
LIME \cite{Ribeiro:2016} is another approach for visualizing neural models (not necessarily textual).
\camready{\citet{,Yogatama:2014} introduced structured sparsity, which encodes linguistic information into the regularization of a model, thus allowing to visualize the contribution of different bag-of-word features.}

Other works jointly learned to encode text and extract the span which best explains the model's prediction \citep{Yessenalina:2010,Lei:2016}.
\citet{Li:2016b} and \citet{Kadar:2017} suggested a method that erases pieces of the text in order to analyze their effect on  a neural model's decisions.
Finally, several works presented methods to visualize deep CNNs \cite{Zeiler:2014,Simonyan:2014,Yosinski:2015},
focusing on visualizing the different layers of the network, mainly in the context of image and video understanding. 
We believe these two types of research approaches are complementary:
inventing general purpose visualization tools for existing black-box models on the one hand, 
and on the other, designing models like \SoftP\ that are interpretable by construction.

\section{Conclusion}

We introduced \SoftP, a novel model that combines neural representation learning with WFSAs.
We showed that \SoftP is \relat a one-layer CNN.
It naturally models flexible-length spans with insertion and deletion, 
and it can be easily customized by swapping in different semirings. 
\SoftP performs on par with or strictly better than four baselines on three text classification tasks, while requiring fewer parameters than the stronger baselines.
On smaller training sets, \SoftP outperforms all four baselines.
As a simple version of an RNN, which is  more expressive than one-layer CNNs, 
we hope that \SoftP will encourage future research on the bridge between these two mechanisms.
To facilitate such research, we release our implementation at \codeurl.

\section*{Acknowledgments}
We thank Dallas Card,  Elizabeth Clark,  Peter Clark,  Bhavana Dalvi,  Jesse Dodge,  Nicholas FitzGerald, Matt Gardner, Yoav Goldberg, Mark Hopkins, Vidur Joshi, Tushar Khot, Kelvin Luu, Mark Neumann,  Hao Peng,  Matthew E.~Peters,  Sasha Rush, Ashish Sabharwal,  Minjoon Seo,  Sofia Serrano, Swabha Swayamdipta,  Chenhao Tan, Niket Tandon,  Trang Tran,  Mark Yatskar,  Scott Yih,  Vicki Zayats, Rowan Zellers, Luke Zettlemoyer, and several anonymous reviewers for their helpful advice and feedback.
This work was supported in part by
NSF grant IIS-1562364, by the Extreme Science and Engineering Discovery Environment (XSEDE),
which is supported by NSF grant ACI-1548562, 
and by the NVIDIA Corporation through the donation of a Tesla GPU.

\bibliography{soft_patterns}
\bibliographystyle{acl_natbib}

\clearpage
\begin{appendices}

\isection{Experimental Setup}{Experimental}

We implemented all neural models in PyTorch,\camready{\footnote{\url{https://pytorch.org/}}}
and the {\bf Hard} baseline in scikit-learn\camready{ \cite{scikit-learn}}.\footnote{\url{http://scikit-learn.org/}}
We train using Adam \cite{Kingma:2014} with a batch size of 150.
We use 300-dimensional GloVe 840B embeddings \cite{Pennington:2014} normalized to unit length.
We randomly initialize all other parameters.
Our MLP has two layers.
For regularization, we use dropout.\footnote{{\bf DAN} uses word dropout instead of regular dropout as its only learnable parameters are the MLP layer weights.}

In all cases, we tune the hyperparameters of our model on the development set by running 30 iterations of random search.
The full list of hyperparameters explored for each model can be found in \tabref{Hyperparameters}.
Finally, we train all models for 250 epochs, stopping early if development loss
does not improve for 30 epochs.

\begin{table}[t]
\ra{1.3}
\scriptsize
\begin{tabularx}{\linewidth}{@{}l X X@{}}
  \toprule
  {\bf Type} & {\bf Values} & {\bf Models} \\
  \midrule
 {\bf Patterns} &
   \{5:10,4:10,3:10,2:10\},
   \{6:10,5:10,4:10\},
   \{6:10,5:10,4:10,3:10,2:10\},
   \{6:20,5:20,4:10,3:10,2:10\},
   \{7:10,6:10,5:10,4:10,3:10,2:10\} &

   {\bf \SoftP} \\ 
 \midrule[\cmidrulewidth]
{\bf Learning rate} & 0.01, 0.05, 0.001, 0.005 & {\bf \SoftP, DAN, BiLSTM, CNN} \\ 
 \midrule[\cmidrulewidth]
{\bf Dropout} & 0, 0.05, 0.1, 0.2 & {\bf \SoftP, BiLSTM, CNN} \\ 
 \midrule[\cmidrulewidth]
{\bf MLP hid.~dim.} & 10, 25, 50, 100, 300 &  {\bf \SoftP, DAN, BiLSTM, CNN} \\ 
 \midrule[\cmidrulewidth]
{\bf Hid.~layer dim.} & 100, 200, 300&  {\bf BiLSTM} \\ 
 \midrule[\cmidrulewidth]
{\bf Out.~layer dim.} & 50, 100, 200&  {\bf CNN} \\ 
 \midrule[\cmidrulewidth]
{\bf Window size} & 4, 5, 6&  {\bf CNN} \\ 
 \midrule[\cmidrulewidth]
{\bf Word dropout} & 0.1, 0.2, 0.3, 0.4 &  {\bf DAN} \\ 
 \midrule[\cmidrulewidth]
{\bf Log.~reg.~param} &  1, 0.5, 0.1, 0.05, 0.01, 0.005, 0.001 &  {\bf Hard} \\ 
 \midrule[\cmidrulewidth]
{\bf Min.~pattern freq.} &  2--10, 0.1\% &  {\bf Hard} \\
  \bottomrule
\end{tabularx}
\caption{\label{Hyperparameters} 
The hyperparameters explored in our experiments.
{\bf Patterns}: the number of patterns of each length.
For example, \{5:20,4:10\} means 20 patterns of length 5 and 10 patterns of length 4.
{\bf MLP hid.~dim.}: the dimension of the hidden layer of the MLP.
{\bf Hid.~layer dim.}: the BiLSTM hidden layer dimension.
{\bf Out.~layer dim.}: the CNN output layer dimension.
{\bf Window size}: the CNN window size.
{\bf Log.~reg.~param}: the logistic regression regularization parameter.
{\bf Min.~pattern freq.}: minimum frequency for a pattern to be included as a logistic regression feature, expressed either as absolute count or as relative frequency in the train set.
{\bf Models}: the models to which each hyperparameter applies (see \secref{Experiments}).
}
\end{table}

\end{appendices}

\end{document}